\documentclass[letterpaper, 10 pt, conference]{ieeeconf}  % Comment this line out if you need a4paper

\IEEEoverridecommandlockouts                              % This command is only needed if 
\usepackage{graphics} % for pdf, bitmapped graphics files
\usepackage{epsfig} % for postscript graphics files
\usepackage{mathptmx} % assumes new font selection scheme installed
\usepackage{times} % assumes new font selection scheme installed
\usepackage{amsmath} % assumes amsmath package installed
\usepackage{amssymb}  % assumes amsmath package installed
\usepackage{mathrsfs}
\usepackage{multirow}
\usepackage[table,xcdraw]{xcolor}
\usepackage{color}
\usepackage{subcaption}
\title{\LARGE \bf
SwinDepth: Unsupervised Depth Estimation using Monocular Sequences \\via Swin Transformer and Densely Cascaded Network
}

\author{Dongseok Shim and H. Jin Kim$^{*}$% <-this % stops a space
\thanks{Authors are with Interdisciplinary Program in Artificial Intelligence, Seoul National University. E-mail: 
        {\tt\small \{tlaehdtjr01, hjinkim\}@snu.ac.kr} $^{*}$Corresponding author}%
}

\begin{document}

\maketitle
\thispagestyle{empty}
\pagestyle{empty}

%%%%%%%%%%%%%%%%%%%%%%%%%%%%%%%%%%%%%%%%%%%%%%%%%%%%%%%%%%%%%%%%%%%%%%%%%%%%%%%%
\begin{abstract}

Monocular depth estimation plays a critical role in various computer vision and robotics applications such as localization, mapping, and 3D object detection. 
Recently, learning-based algorithms achieve huge success in depth estimation by training models with a large amount of data in a supervised manner.
However, it is challenging to acquire dense ground truth depth labels for supervised training, and the unsupervised depth estimation using monocular sequences emerges as a promising alternative. 
Unfortunately, most studies on unsupervised depth estimation explore loss functions or occlusion masks, and there is little change in model architecture in that ConvNet-based encoder-decoder structure becomes a de-facto standard for depth estimation. 
In this paper, we employ a convolution-free Swin Transformer as an image feature extractor so that the network can capture both local geometric features and global semantic features for depth estimation.
Also, we propose a Densely Cascaded Multi-scale Network (DCMNet) that connects every feature map directly with another from different scales via a top-down cascade pathway. 
This densely cascaded connectivity reinforces the interconnection between decoding layers and produces high-quality multi-scale depth outputs. 
The experiments on two different datasets, KITTI and Make3D, demonstrate that our proposed method outperforms existing state-of-the-art unsupervised algorithms.
\end{abstract}

\section{introduction}
Estimating a high-quality depth map from a single RGB image is theoretically an ill-posed problem without additional ques for triangulation. 
Learning-based methods aims to mitigate such difficulties by training a neural network in a supervised manner with a large amount of RGB images and densely annotated depth labels \cite{eigen2014depth, laina2016deeper, fu2018deep}. 
However, it is still challenging to collect accurate ground truth depth for supervised learning and, as an alternative, demands for unsupervised training of depth estimation network increase.
There are two promising approaches to train depth estimation networks without depth labels which use stereo-paired images \cite{garg2016unsupervised, godard2017unsupervised} or monocular image sequences \cite{zhou2017unsupervised, godard2019digging}. In particular, the latter one is more attractive in that it does not require additional sensors such as a stereo camera. 

\begin{figure}[t]
    \centering
    \includegraphics[width=0.48\textwidth]{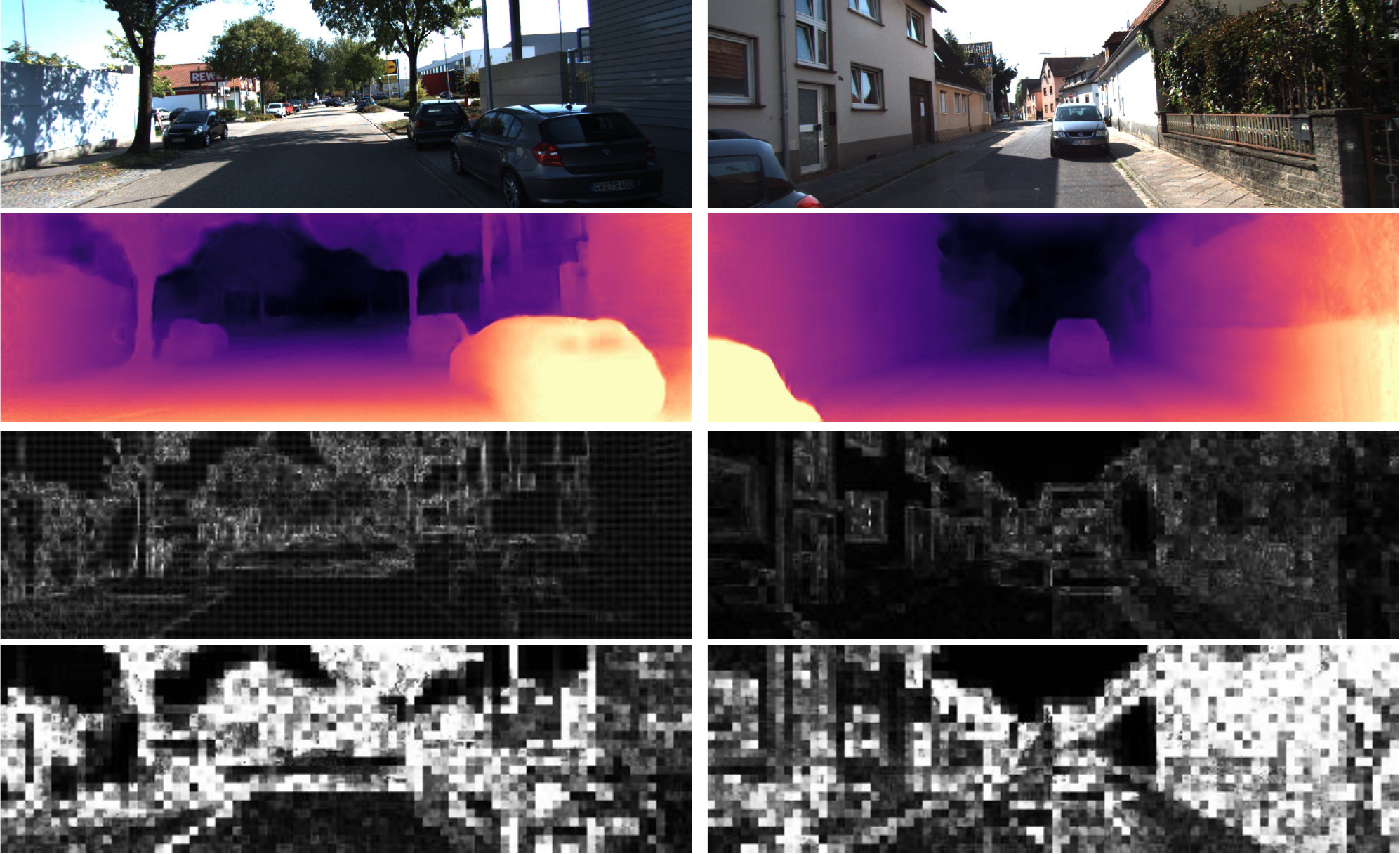}
    \caption{A single image depth estimation with a hierarchical Transformer and densely cascaded network. From top to bottom, each row indicates input RGB images, estimated depth maps, and self-attention masks from multiple heads in Transformer feature extractor.}
    \vspace{-10pt}
    \label{fig:my_label}
\end{figure}

Training only with monocular sequences needs to jointly estimate the depth map and the ego-motion of the camera to calculate the transformation matrix for photometric loss that it requires an additional pose estimation network.
Also, it assumes a static scene and a moving camera for rigid motion. 
Due to these assumptions, the estimation errors such as undesired holes appear in the depth map if there exist some dynamic objects, e.g. pedestrians or cars, which can be a major hurdel for robotics applications \cite{cadena2016past, yu2018ds}.
%Also, it assumes a static scene and rigid motion that the estimation errors such as undesired holes appear in the depth map if there exist some dynamic objects, e.g. pedestrians or cars, which leads to non-rigid motion. 

To overcome these errors, follow-up studies redesign the photometric loss function or propose masking strategies to prevent occlusions and hide moving objects in the scene, which may improve the quality of estimated depth maps.
Unfortunately, there are few studies to improve the architecture of the depth estimation network itself where U-Net structure \cite{ronneberger2015u} with Convolutional neural Network (ConvNet) -based feature extractor \cite{simonyan2015deep, szegedy2015going, he2016deep} is dominant both in supervised and unsupervised training.

Recently, Vision Transformer (ViT) \cite{dosovitskiy2020image} shows that a pure Transformer produces better performance in image classification compared to competitive ConvNet. Its follow-up studies also show that ViT is still effective for backbone architecture in various visual recognition tasks such as object detection \cite{carion2020end, beal2020transformerbased} and semantic segmentation \cite{zheng2021rethinking}.

There exist some approaches trying to apply ViT on
depth estimation, but most of them use feature maps generated by  ConvNet \cite{bhat2021adabins, li2021revisiting}, rather than directly using RGB image as the input of Transformer.
Some research such as DPT \cite{ranftl2021vision} na\"ively stack multiple ViTs to generate ConvNet-like hierarchical feature maps to leverage Transformer-based encoder for downstream networks with skip connection, but it leads to heavy computational cost and many model parameters.

In this paper, we utilize a convolution-free hierarchical Transformer with a downsizing module and attention window from Swin Transformer \cite{liu2021swin} to reduce the innate computational burden in calculating feature attention. 
The resolution of the feature map progressively gets smaller by merging nearby image patches in deeper layers, and the attention map is calculated in the fixed-size non-overlapping windows. 
These merits make the hierarchical Transformer encoder become much lighter while keeping the strength of capturing semi-global image features. It produces better performance on depth estimation compared to existing ConvNet-based architecture that has twice as many model parameters.

% So, the difference between our approach and previous Transformer-based depth estimation algorithms are (1) we leverage window-based attention and downsizing module in Transformer for computation efficiency, and (2) our method do not use any ConvNet for feature extraction whereas other Transformer-based depth estimation takes advantage of ConvNet feature extractors so-called hybrid ViT.

Also, we propose a simple but effective feature aggregation model, Densely Cascaded Multi-scale Network (DCMNet), to estimate multi-scale depth maps using hierarchical image features from Transformer encoder. 
Inspired by Dense Convolutional Network (DenseNet) \cite{huang2017densely} and Feature Pyramid Network (FPN) \cite{lin2017feature}, we leverage the dense concatenation and the top-down cascade addition to improve the interconnection between features from decoding layers. We densely connect the hierarchical feature maps directly with one another that all the coarse low-dimension features are exploited to generate fine high-resolution depth outputs without any attenuation by passing through decoding layers.

In short, the contributions of our paper can be summarized as follows:
\begin{itemize}
\item We show that pure Transformer-based hierarchical feature extractor outperforms competitive ConvNet-based architecture in depth estimation.
\item We propose DCMNet which improves the interconnection between decoding layers with a top-down cascade pathway and dense concatenation to produce high-quality multi-resolution depth outputs.
\item We demonstrate the effectiveness of our approach on KITTI \cite{geiger2012we} and Make3D dataset \cite{saxena2008make3d} compared to existing state-of-the-art unsupervised depth estimation. 
\end{itemize}
% The official implementation code of the paper is provided in supplementary materials and will be publicly available after camera-ready submission.
\section{Related Work}
In this section, we summarize the previous research and the applications on Vision Transformer, and review learning-based monocular depth estimation algorithms which are categorized into supervised and unsupervised approaches. 
\subsection{Vision Transformer}
Transformer architecture with self-attention \cite{vaswani2017attention} first suggested for machine translation has become one of the most widely used models in natural language processing (NLP).

To leverage the advantage of Transformer in NLP, Vision Transformer (ViT) \cite{dosovitskiy2020image} divides an RGB image into several patches and flatten them to treat the image as sequential data for Transformer input.
Convolution-free Transformer encoder with patch embedding achieves a huge success and outperforms state-of-the-art ConvNets in image classification. Later, DeiT \cite{touvron2021training} introduces an additional distillation token to alleviate the dependency on the large-scale dataset for pre-training.

SETR \cite{zheng2021rethinking} utilizes ViT as the backbone network and the feature aggregation module for semantic segmentation. DPT \cite{ranftl2021vision} stacks multiple ViTs to generate hierarchical feature maps for dense prediction.
As stacking ViT modules without downsizing makes the network too large, Swin Transformer \cite{liu2021swin} proposes a patch merging and attention window to reduce the computational and memory burden of Transformer block. It produces better performance as a backbone architecture for image classification, object detection and semantic segmentation compared to competitive ConvNet architecture, ViT and its variants.

In this paper, we also utilize the patch merging strategy and attention window to formulate a hierarchical Transformer feature extractor, hypothesizing that they will provide the same advantage shown in other visual recognition tasks for unsupervised depth estimation as well.
\subsection{Learning-based Depth Estimation}
It is challenging to generate high-quality depth outputs with a single RGB image and the development of deep learning relaxes difficulties in depth estimation with deep neural networks and a large-scale training dataset.
%and the advent of deep learning relaxes such difficulties with a large-scale training dataset.

Eigen \cite{eigen2014depth} proposes an end-to-end supervised training algorithm which consists of two-scale ConvNets each generating low-dimension coarse and high-dimension fine depth outputs.
Laina \cite{laina2016deeper} utilizes ResNet \cite{he2016deep} with its novel up-projection network for fast and efficient training, and DORN \cite{fu2018deep} formulates depth estimation as a discrete problem by proposing an ordinal regression loss.

Due to the poor availability of depth labels for supervised learning, 
%studies for training a depth estimation network in an unsupervised manner appear especially using monocular consecutive frames.
unsupervised depth estimation emerges as a promising alternative especially using consecutive monocular frames.
%\citet{garg2016unsupervised} propose an unsupervised depth estimation framework that exploits synchronized stereo images and pixel-wise reprojection loss to implicitly generate depth maps without any supervision. 
%\citet{godard2017unsupervised} additionally adopt left-right consistency loss that the depth estimation network generates consistent outputs both in left and right images.
%Still, these methods rely on stereo-paired images which require an additional stereo camera, and it motivates the development of depth estimation using only monocular consecutive frames.
SfMLearner \cite{zhou2017unsupervised} jointly estimates depth and the ego-motion of the camera with two different networks. 
The photometric loss is calculated by projecting one monocular frame to the other camera coordinate via Transformation matrix from the outputs of the pose estimation network and estimated depth values.
The problem is that it assumes a rigid motion and the presence of moving objects leads to poor performance both in pose and depth estimation.
To alleviate the performance degradation, 
%SfMLearner \cite{zhou2017unsupervised} introduces a predictive mask to learn undesirable non-rigid regions and GeoNet \cite{yin2018geonet} decouples the depth map and optical flow that it considers the object motion in image sequences. 
Monodepth2 \cite{godard2019digging} leverages a minimum operation in calculating the photometric reprojection loss to prevent occlusion and proposes an auto-generated mask to hide moving objects in the scene. 

We adopt the unsupervised depth estimation strategy suggested in Monodepth2 to estimate both depth values and relative pose with monocular image sequences.
\section{Method}
\begin{figure*}[t]
    \centering
    \includegraphics[width=0.8\textwidth]{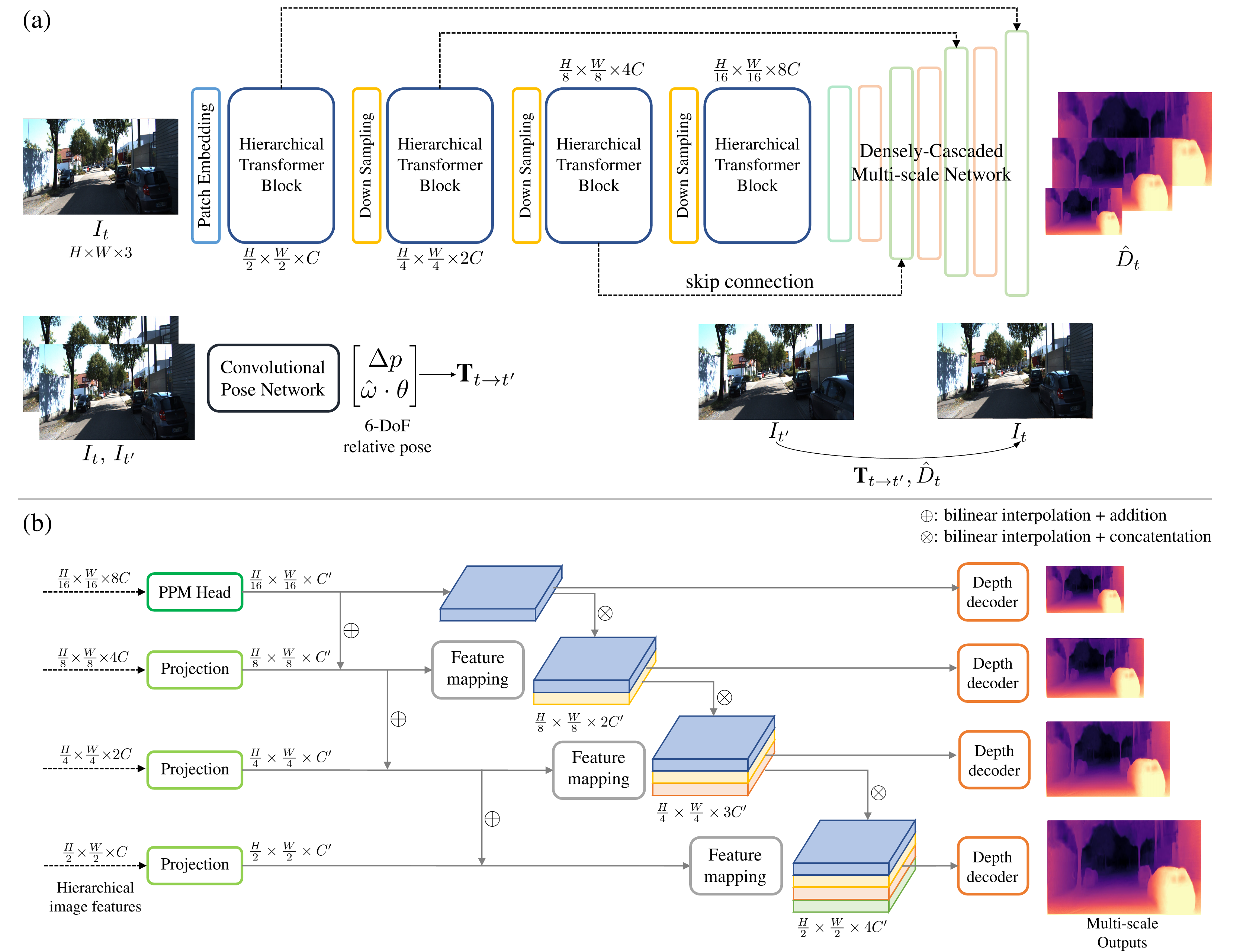}
    \caption{(a) Overall training algorithm of unsupervised depth estimation with the hierarchical Transformer-based encoder and DCMNet. The depth estimation network consists of the Transformer encoder followed by DCMNet and the pose estimation network consists of ConvNet with 6-dimension outputs. (b) Illustration of the architectural details of DCMNet. Top-down cascade pathway reinforces interconnection between decoding layers with element-wise addition and dense concatenation.}
    \label{fig:overall}
\end{figure*}

In this section, we elaborate the architectural details and key ideas of hierarchical Transformer and our proposed Densely Cascaded Multi-scale Network (DCMNet). 
Then, we describe the training algorithm for unsupervised depth estimation using monocular image sequences. 
An overview of our depth estimation network and the training algorithm are depicted in Figure \ref{fig:overall}.

\subsection{Architecture}
\subsubsection{Hierarchical Transformer}
Similar to Vision Transformer (ViT) \cite{dosovitskiy2020image}, a hierarchical Transformer first splits the input RGB image into several non-overlapping patches ($2\times 2$). 
Image patches are then flattened and mapped to $C$-dimensional feature space by trainable linear projection in the patch embedding module.

We adopt some architecture designs suggested in Swin Transformer \cite{liu2021swin} to formulate the convolution-free hierarchical Transformer for depth estimation.
The difference between ViT and the hierarchical Transformer is that the hierarchical Transformer produces multiple feature maps with different resolutions while ViT produces a single feature map with the same resolution of the input feature.
The feature maps of hierarchical Transformer progressively get smaller that the feature resolution is halved after passing through a downsampling module. 
We utilize a patch merging strategy \cite{liu2021swin} for the downsampling module to concatenate each group of 2$\times$2 neighboring patches in the feature map. 
It leads to 2$\times$ downsizing in the spatial size and 4$\times$ expansion in the feature dimension. 
Following, a linear layer is applied to the 4$C$-dimension merged features and projects them to 2$C$-dimensional space. 
We adopt 4 hierarchical Transformer blocks, and feature maps with 4 different resolutions are produced (1/2, 1/4, 1/8, 1/16). 

The other difference is that the hierarchical Transformer replaces the standard multi-head self-attention (MSA) with window-based self-attention (W-MSA) \cite{liu2021swin} to reduce the innate computation complexity in calculating attention. 
As global self-attention requires a quadratic complexity with respect to the input resolution, it only calculates local attention on the non-overlapping window with fixed size $w$, and the computation reduces to the linear complexity.

A sole window-based self-attention relaxes computational costs but has difficulty in capturing global features due to the limited connection between windows.
To overcome these lack connections, we also leverage shifted window self-attention (SW-MSA) \cite{liu2021swin} and reinforce cross-window connections. 
The shifted window formulates its attention window by translating the window partitioning by the half size of regular window, $\frac{w}{2} \times \frac{w}{2}$, from the top-left pixel. 
Then, it calculates self-attention on the newly shaped window configuration with different sizes.

A single hierarchical block consists of W-MSA, SW-MSA and MLP modules. LayerNorm (LN) layer is applied before each module, and residual connections after each module. 
MLP module contains 2-layer with a GELU non-linear activation function. 
Multi-resolutional feature maps are generated by computing consecutive hierarchical Transformer blocks as 
\begin{align}
    &{{\bf{z}^{\prime}}_{l}} = \text{W-MSA}\left( {\text{LN}\left( {{{\bf{z}}_{l - 1}}} \right)} \right) + {\bf{z}}_{l - 1},\nonumber\\
    &{{\bf{z}}_l} = \text{MLP}\left( {\text{LN}\left( {{{\bf{z}^{\prime}}_{l}}} \right)} \right) + {{\bf{z}^{\prime}}_{l}},\nonumber\\
    &{{\bf{z}^{\prime}}_{l+1}} = \text{SW-MSA}\left( {\text{LN}\left( {{{\bf{z}}_{l}}} \right)} \right) + {\bf{z}}_{l}, \nonumber\\
    &{{\bf{z}}_{l+1}} = \text{MLP}\left( {\text{LN}\left( {{{\bf{z}^{\prime}}_{l+1}}} \right)} \right) + {{\bf{z}^{\prime}}_{l+1}}, \label{eq.swin}
\end{align}
where ${\bf{z}^{\prime}}_{l}$ and ${\bf{z}}_{l}$ denote the features from (S)W-MSA module and the MLP module in the $l^{th}$ layer of Transformer block respectively.

Thanks to these advantages of hierarchical Transformer, the depth estimation network can leverage hierarchical feature maps to generate dense predictions via skip connection strategy with low computational costs and model parameters.

\begin{table*}[t!]
\caption{Quantitative results. Comparison of our proposed method to existing methods on KITTI 2015 \cite{geiger2012we} using train/test split in Eigen \cite{eigen2014depth}. 
Best results are presented in \textbf{bold} and the second best are \underline{underlined}. 
All the results are presented without post-processing suggested in \cite{godard2017unsupervised}. Train column indicates available data during training as; D - Depth supervision, $\text{D}^{*}$ - Auxiliary depth supervision, M - Monocular self-supervision, S - Stereo self-supervision and \textdagger\, indicates newer results from github. 
R18 and R50 denote ResNet-18 and ResNet-50 backbone respectively. 
%The red metrics indicate \textit{lower is better} and blue metrics indicate \textit{higher is better}.
The error metrics (Abs Rel, Sq Rel, RSME, RMSE log) indicate \textit{lower is better} and the accuracy metrics ($\delta<1.25^{i}$) indicate \textit{higher is better}.
}
 \centering
\resizebox{0.9 \textwidth}{!}{
  \begin{tabular}{|l|c||c|c|c|c|c|c|c|}
  \hline
  Method & Train & Abs Rel & Sq Rel & RMSE  & RMSE log & $\delta < 1.25 $ & $\delta < 1.25^{2}$ & $\delta < 1.25^{3}$\\
  \hline
Eigen \cite{eigen2014depth} & D & 0.203 & 1.548 & 6.307 & 0.282 & 0.702 & 0.890 & 0.890\\
Liu \cite{liu2015learning}  & D & 0.201 & 1.584 & 6.471 & 0.273 & 0.680 & 0.898 & 0.967\\
Klodt \cite{klodt2018supervising}  & D*M & 0.166 & 1.490 & 5.998 & - &  0.778 & 0.919 & 0.966\\
AdaDepth \cite{kundu2018adadepth}   & D* & 0.167 & 1.257 & 5.578 & 0.237 & 0.771 & 0.922 & 0.971\\
Kuznietsov \cite{kuznietsov2017semi}  & DS & 0.113 & 0.741 & 4.621 & 0.189 & 0.862 & 0.960 & 0.986\\
DVSO \cite{yang2018deep}  & D*S & 0.097 & 0.734 & 4.442 & 0.187 & 0.888 & 0.958 & 0.980\\
SVSM FT\cite{luo2018single}  & DS & \underline{0.094} & \underline{0.626} & 4.252 & 0.177 & 0.891 & 0.965 & 0.984\\
Guo \cite{guo2018learning}  & DS & 0.096 & 0.641  & \underline{4.095}  & \underline{0.168}  & \underline{0.892}  & \underline{0.967}  & \underline{0.986} \\
DORN \cite{fu2018deep}  & D & \textbf{0.072}&  \textbf{0.307} & \textbf{2.727} & \textbf{0.120} & \textbf{0.932} & \textbf{0.984} & \textbf{0.994}\\ 

\hline

SfMLearner \cite{zhou2017unsupervised} \textdagger & M & 0.183 & 1.595 & 6.709 & 0.270 & 0.734 & 0.902 & 0.959\\
Yang \cite{yang2018unsupervised}  & M & 0.182 & 1.481  & 6.501  & 0.267  & 0.725  & 0.906  & 0.963\\
Mahjourian \cite{mahjourian2018unsupervised}  & M & 0.163 & 1.240 & 6.220 & 0.250 & 0.762 & 0.916 & 0.968\\

GeoNet \cite{yin2018geonet} \textdagger & M  & 0.149 & 1.060 & 5.567 & 0.226 & 0.796 & 0.935 & 0.975\\
DDVO \cite{wang2018learning}  & M  & 0.151 & 1.257 & 5.583 & 0.228 & 0.810 & 0.936 & 0.974\\
DF-Net \cite{zou2018df} & M & 0.150 & 1.124 & 5.507 & 0.223 & 0.806 & 0.933 & 0.973\\
LEGO \cite{yang2018lego} & M & 0.162 & 1.352 & 6.276 & 0.252 & - & - & - \\
Ranjan \cite{ranjan2019competitive}  & M & 0.148 & 1.149 & 5.464 & 0.226 & 0.815 & 0.935 & 0.973\\
EPC++ \cite{luo2019every}  & M & 0.141 & 1.029 & 5.350 & 0.216 & 0.816 & 0.941 & 0.976\\
Struct2depth \cite{casser2019depth}  & M & 0.141 & 1.026 & 5.291 &  0.215 & 0.816 & 0.945 & 0.979\\

Monodepth2 \cite{godard2019digging} R18 & M &   
 0.115 &   0.903 &   4.863 &   0.193 &   0.877 &   0.959 &   0.981 \\ 

Monodepth2 \cite{godard2019digging} R50 & M &   
 0.110 &   0.831 &   4.642 &   0.187 &   0.883 &   \underline{ 0.962} &   0.982 \\ 
 
 PackNet-SfM \cite{guizilini20203d}& M &   
 0.111 &   \underline{ 0.785} &   \underline{ 4.601} &   0.189 &   0.878 &   0.960 & 0.982 \\ 
 
 Lyu \cite{lyu2021hr}& M &   
 \underline{0.109} &   0.792 &   4.632 &   \underline{ 0.185} &   \underline{ 0.884} &   \underline{ 0.962} &   \underline {0.983} \\

\hline
\bf{Ours}&M& {\bf 0.106} &   {\bf 0.739} &   {\bf 4.510} &   {\bf 0.182} &   {\bf 0.890} &   {\bf 0.964} &   {\bf 0.984}\\

\hline
  \end{tabular}
}
\vspace{-10pt}
%\vspace{-5pt}
\label{tab:quan}
\end{table*}

\subsubsection{Densely Cascaded Multi-scale Network}
Existing depth estimation networks mostly adopt U-Net style depth decoder \cite{ronneberger2015u} which utilizes the ConvNet followed by bilinear interpolation and improve the connection with the hierarchical encoder via skip connection. 
The problem is that the sole skip connection cannot improve the connectivity between the features inside the depth decoder. 
To overcome this limited connection, we leverage the dense concatenation \cite{huang2017densely} and the top-down addition \cite{lin2017feature} to reinforce the interconnection of decoding layers, simultaneously taking advantage of the improved connection between the encoder and the depth decoder using skip connection.

Hierarchical image features of 4 different resolutions (1/2, 1/4, 1/8, 1/16) from the encoder are passed to each stage of the decoding network. 
Each feature is first projected to the same $C^{\prime}$-dimensional space and the features from different stages are subsequently connected by element-wise addition followed by bilinear interpolation via top-down \textit{cascade} pathway,
\begin{equation}
    \textbf{x}_{s} = \sum_{k=1}^{s} (\textbf{f}_{k})_{\uparrow}
\end{equation}
where $\textbf{f}_{k}$ and $(\cdot)_{\uparrow}$ denote projected features from $k^{\text{th}}$ stage and bilinear interpolation respectively.
The connected features are mapped to the latent space with the same dimension $C^{\prime}$ with a non-linear feature mapping module in order to alleviate the aliasing effect by bilinear upsampling.
For much denser connectivity in the decoder, we directly connect all the subsequent features from the feature mapping module $\bf{x}_{1}^{\prime}, ..., \bf{x}_{s}^{\prime}$ by concatenation to mitigate the attenuation of the low-resolution image features by passing through deep decoding layers. 
Consequently, the depth decoder of the ${s}^{\text{th}}$ cascade stage $H_{s}(\cdot)$ receives the feature maps of all the preceding stages as the input:
\begin{equation}
    \hat{D}_{s} = H_{s}([\bf{x}_{1}^{\prime}, ..., \bf{x}_{s}^{\prime}]_{\uparrow\text{cat}})
\end{equation}
where $[\,\cdot\,]_{\uparrow\text{cat}}$ and $\hat{D}_{s}$ denote the channel-wise concatenation of all the elements after bilinear interpolation and the estimated depth map in the $s^{\text{th}}$ stage respectively. 
The estimated depth maps have different spatial size according to the cascade stage of DCMNet so that each stage produces depth outputs with different ratios (1/1, 1/2, 1/4, 1/8) of the input resolution.

Additionally, we employ a Pyramid Pooling Module (PPM) from PSPNet \cite{zhao2017pyramid} on the first cascade stage and replace it with the regular projection module. As the hierarchical Transformer backbone cannot capture the fully-global representation, the PPM head helps the encoder to understand the entire scene information by providing a global contextual prior.
It applies multiple average pooling with different sizes (1, 2, 3, 6) after the last layer of the Transformer encoder that it brings effective global prior representations for downstream depth estimation decoder.

% The depth decoder leverages both $3\times 3$ and $1\times 1$ convolution layers followed by ReLU and Sigmoid non-linearity respectively.
% The projection module and the feature mapping module utilize a single $3\times 3$ and $1\times 1$ convolution layer followed by ReLU non-linearity respectively. 
% Dense features are 2$\times$ upsampled before the first layer of the depth decoder.
% The architecture of DCMNet is illustrated in Figure \ref{fig:overall}(b).

The depth decoder leverages both $3\times 3$ and $1\times 1$ convolution layers, whereas
the projection module and the feature mapping module respectively utilize a single $3\times 3$ and $1\times 1$ convolution layer. 
Dense features are 2$\times$ upsampled before the first layer of the depth decoder,  and no convolution operation changes the spatial size of the features.
The architecture of DCMNet is illustrated in Figure \ref{fig:overall}(b).

\subsection{Unsupervised Depth Estimation}
We use monocular image sequences to train a depth estimation network in an unsupervised manner. We jointly train depth estimation and pose estimation network to predict the depth map of the image and the ego-motion of the camera.
We build a depth estimation network as a hierarchical Transformer followed by DCMNet and pose estimation networks as ResNet-18 \cite{he2016deep} with 6-dimension outputs.
As our depth estimation network estimates multi-scale depth maps, we explain the training algorithm with a single scale depth map and its corresponding resized target image $I_{t}$ for simplicity. The source image $I_{t^{\prime}}$ for estimating ego-motion is sampled from the temporally adjacent frames of the monocular video $t^{\prime} \in \{t-1, t+1\}$.

\section{Experiment}
In this section, we demonstrate the effectiveness of our proposed method on KITTI 2015 \cite{geiger2012we} and Make3D dataset \cite{saxena2008make3d}. We also validate that (1) the hierarchical Transformer outperforms depth estimation performance compared to previous ResNet \cite{he2016deep} and ViT/DeiT \cite{dosovitskiy2020image, touvron2021training} backbone with much smaller model parameters, and (2) how each component of DCMNet contributes to the network performance.
\subsection{Dataset}
We train and evaluate our method on KITTI with train/test data split suggested in Eigen \cite{eigen2014depth}. 
Following SfMLearner \cite{zhou2017unsupervised} and Monodepth2 \cite{godard2019digging}, we remove static frames in monocular sequences, which results in 39,810 images for training and 4,424 images for validation. 
The resolution of the image is set to 640$\times$192, which is most widely adopted to evaluate depth estimation performance on KITTI dataset. 
We also evaluate our method on 134 test images from Make3D dataset without any fine-tuning.

\subsection{Implementation Details}
We implement our methods on the public deep learning platform PyTorch \cite{paszke2017automatic} and train them on 4 Nvidia RTX 3090 GPUs. 
We adopt Adam \cite{kingma2017adam} optimizer with $\beta_{1}=0.9$, $\beta_{2}=0.999$, and a batch size of 12 for 40 epochs. We set the initial learning rate as $10^{-4}$ which is decayed after the first 15 epochs by factor of 10. 
Similar to \cite{guo2018learning, godard2019digging}, we train the model with weights pre-trained on ImageNet-1k \cite{ILSVRC15} to reduce the training time and improve the overall model performance.

We build the hierarchical Transformer and DCMNet to obtain the model size similar to or smaller than the existing unsupervised depth estimation network. We set the window size of the hierarchical Transformer $w$ as 4 and the dimension of the hidden layers in the first block $C$ as 64. The projected dimension $C^{\prime}$ of DCMNet is set to 128.
\subsection{KITTI Results}

\begin{figure*}[t]
    \centering
    \includegraphics[width=1.0 \textwidth]{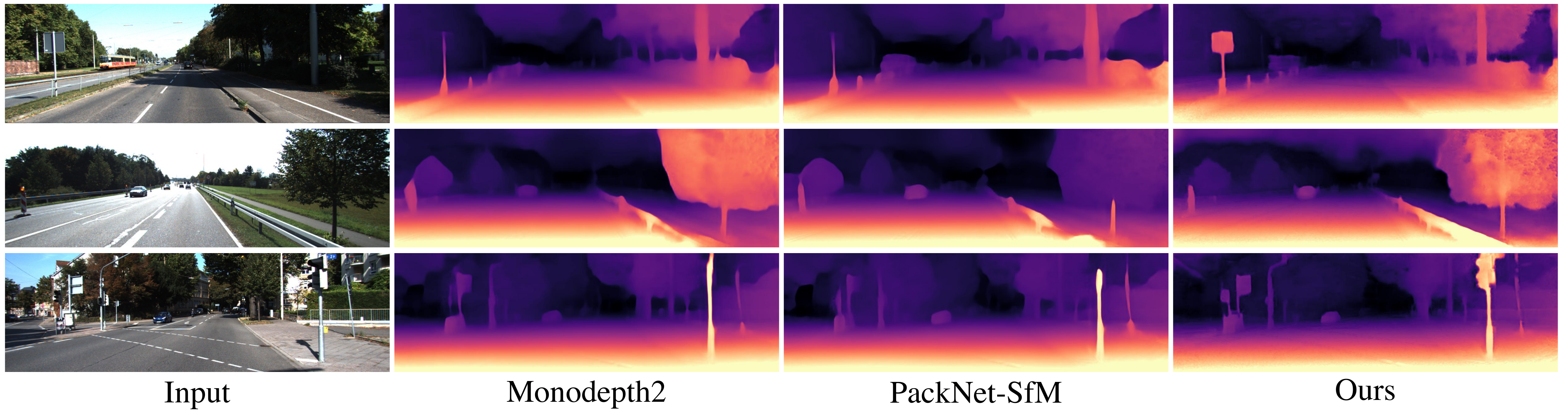}
    \caption{Qualitative results on KITTI. Our method performs better on small objects such as traffic lights, signs, and guardrails compared to competitive unsupervised algorithms; Monodepth2 \cite{godard2019digging} and PackNet-SfM \cite{guizilini20203d}.}
    \vspace{-10pt}
    \label{fig:qual}
\end{figure*}

We compare our model with other competitive depth estimation algorithms on KITTI 2015 with the image resolution of $640\times192$ at maximum using both standard error and accuracy metrics. 
During evaluation, depths are capped to 80 m and per-image median ground truth scaling \cite{zhou2017unsupervised} is applied for monocular self-supervised methods. 
As shown in Table \ref{tab:quan}, our proposed method outperforms existing state-of-the-art unsupervised approaches. 
Especially, our method achieves better results compared to recent larger networks, i.e. Monodepth2 \cite{godard2019digging} with ResNet-50 backbone (35M) and PackNet-SfM \cite{guizilini20203d} (120M), by leveraging relatively small model parameters (25M). 
Qualitative results are shown in Figure \ref{fig:qual} and it demonstrates that our method produces distinct boundaries between objects and background compared to blurry edges shown in competitive unsupervised methods.
The estimated depth maps from our methods are capable of recognizing thin, small, but important objects for autonomous driving such as traffic lights, signals and the guardrail, simultaneously recovering the details of the foliage of roadside trees.
We attribute such advantages to the window attention-based semi-global feature mapping by hierarchical Transformer which captures both local geometric and global semantic information of the environment.
\begin{table*}[t!]
      \caption{
      Ablation.
      A comparison between different variants of our model with monocular self-supervision on KITTI 2015 \cite{geiger2012we} using the Eigen split.
      (a) The baseline model is U-Net and replace its backbone with ConvNet-based, ViT-based networks, and Swin Transformer.
      (b) Three components of DCMNet (PPM, Top-down Addition, and Dense Concatenation) are ablated one by one to validate the effectiveness of each component.
      (c) Our SwinDepth produces on-par performance compared to recent depth estimation using hierarchical Transformers. * indicates no ImageNet pre-training as its pre-trained weights are not provided in the official repository.
      }
  \centering
  \resizebox{1.0 \textwidth}{!}
{
    \footnotesize
    \begin{tabular}{|l|l|c||c|c|c|c||c|c|c|c|c|c|c|}
      \hline
      &Method &
      Param &

      \begin{tabular}{@{}c@{}}Skip \\ Connect\end{tabular} &
      PPM&
      \begin{tabular}{@{}c@{}}Top-down \\ Addition\end{tabular} &
      \begin{tabular}{@{}c@{}}Dense \\ Concat\end{tabular} &

    Abs Rel & Sq Rel & RMSE  &
      \begin{tabular}{@{}c@{}}RMSE \\ log\end{tabular} & 
      $\delta<$1.25 & $\delta<$1.25$^{2}$ & $\delta<$1.25$^{3}$ \\
      
      \hline

      (a) &  ResNet-18 + U-Net   & 14.8M  &  \checkmark &  &  &         &  
          0.115 &   0.903 &   4.863 &   0.193 &   0.877 &   0.959 &   0.981 \\  %
        % & ResNet-50 + U-Net& 34.6M & \checkmark &     &  &  & 
        %     \textbf{0.110}  &   0.831  &   \textbf{4.642}  &   \textbf{0.187}  &   0.883  &   0.962  &   \textbf{0.982}  \\
        % \clinegrayone
            & ResNet-101 Dilated + U-Net &51.6M & \checkmark&  &  & 
        &   \textbf{0.110}  &   0.876  &   4.853  &   \textbf{0.189}  &   0.879  &   0.961  &   \textbf{0.982}  \\

          & ViT-B/16 + U-Net & 88.9M & N/A&   &  &  & 
            0.132  &  1.298  &  5.494  &   0.216  & 0.851  & 0.947  &   0.974  \\
    & DeiT-B + U-Net & 88.9M &  N/A&  &  &  &   
      0.127 &   1.126 &   5.313 &   0.210 &   0.857 &   0.949 &   0.976 \\ %

      & Swin + U-Net & 23.0M & \checkmark & &  &  & 
      0.111 &   \textbf{0.864} &   \textbf{4.671} &   \textbf{0.189} &   \textbf{0.887} &   \textbf{0.962} &   \textbf{0.982} \\  %

      \hline %

      (b) 
      & Swin + DCMNet w/o  PPM    &  23.9M &\checkmark &  & \checkmark & \checkmark & 
        0.108 &   0.795 &   4.554 &   0.184 &   \textbf{0.890} &   \textbf{0.964} &   0.983 \\ %

      & Swin + DCMNet w/o TA     & 25.2M &\checkmark & \checkmark&  &  \checkmark& 
        0.107 &   0.747 &   4.551 &   0.183 &   0.888 & \textbf{0.964} &   \textbf{0.984} \\ %

       & Swin + DCMNet w/o DC    & 24.3M &\checkmark & \checkmark & \checkmark &  & 
        0.107 &   0.763 &   4.592 &   0.184 &   0.888 &   0.963 &   0.983 \\ %

       & Swin + DCMNet (full)& 25.2M & \checkmark&  \checkmark& \checkmark & \checkmark & 
        \textbf{0.106} &   \textbf{0.739} &  \textbf{4.510} &   \textbf{0.182} &   
        \textbf{0.890} &   \textbf{0.964} &   \textbf{0.984}\\ %
      \hline %
      (c)& DPT$^{*}$ \cite{ranftl2021vision} & 123M  &&&&& 0.210 &   0.847 & 4.554 & 0.184 &   0.847 &  0.948  &   0.976 \\ %
        & HRFormer \cite{yuan2021hrformer}   & 56.0M &&&&& 0.108 &   0.812 & 4.634 & 0.185 &   0.891 &  0.963  &   0.982 \\ %
    \hline
      \end{tabular}
  }
\label{tab:ablation}
\end{table*}

\begin{table}[t]
    \caption{{Make3D results.} All results of the monocular self-supervision (M) benefit from median-scaling.}
    \centering
    \resizebox{0.98 \linewidth}{!}{
      \begin{tabular}{|l|c||c|c|c|c|}
      \hline
       & Type & Abs Rel & Sq Rel  &RMSE & $\text{log}_{10}$ \\
      \hline
      Karsch \cite{karsch2014depth} & D & 0.428 & 5.079 & 8.389 & 0.149 \\
      Liu \cite{liu2014discrete}& D & 0.475 & 6.562 & 10.05 & 0.165 \\
      Laina \cite{laina2016deeper}& D & {\bf 0.204} & {\bf 1.840} & {\bf 5.683} & {\bf 0.084} \\ \hline
      Monodepth \cite{godard2017unsupervised} & S & 0.544 & 10.94 & 11.760 & 0.193 \\
      SfMLearner \cite{zhou2017unsupervised} & M &  0.383 & 5.321 & 10.470 & 0.478 \\
      DDVO \cite{wang2018learning} & M & 0.387 & 4.720 & 8.090 & 0.204 \\
       Monodepth2 \cite{godard2019digging} & M & 0.322 & 3.589& 7.417 & 0.163  \\
       %\citet{guizilini20203d} & M & 0.324 & 3.443& 7.305 & 0.164  %\\
        \hline
        \textbf{Ours} & M & \textbf{0.295} & \textbf{3.042} & \textbf{6.917} & \textbf{0.150}  \\
    \hline

      \end{tabular}
    }
    \vspace{-10pt}
    \label{tab:make3d}
\end{table}
\subsection{Ablation Study}
For better understanding of our proposed methods, we perform an ablation study in Table \ref{tab:ablation} by changing (a) the feature extractor and (b) components of the dense decoder of the depth estimation network.
We also compare our method with other hierarchical Transformers (c) for dense prediction.

To explore the effectiveness of the feature extractor in the depth estimation, we compare the Swin Transformer-based hierarchical feature extractor (Swin) with the standard ConvNets, i.e. ResNet, and the previous state-of-the-art Transformer networks, i.e. ViT and DeiT, in Table \ref{tab:ablation}(a). We employ U-Net \cite{ronneberger2015u} as the baseline architecture and the evaluation is done by replacing only the backbones with all the network kept the same.
The feature extractors are pre-trained on ImageNet-1k except for ViT-B/16; ViT-B/16 is pre-trained on ImageNet-21k \cite{ridnik2021imagenet21k} because it cannot produce high-quality outputs without pre-training on a large-scale dataset.

Swin Transformer-based hierarchical Transformer achieves the best results in most of the metrics compared to other backbones, including both ConvNet-based and Transformer-based, where ResNet-101 with the dilated convolution \cite{chen2017rethinking} leverages model parameters in double, and ViT-B/16 and DeiT-B in quadruple. 

ViT-B/16 and DeiT-B cannot take advantage of the skip connection strategy as they do not produce hierarchical feature maps, rather they produce a single resolution feature output.
Both show the worst results with the largest model parameters, and it indicates the importance of generating hierarchical feature maps and skip connection in determining the depth estimation performance.

To evaluate how each part of DCMNet contributes to the model performance, we remove three components of the network one by one and report the performance of each ablated model in Table \ref{tab:ablation}(b).
We observe that the ablation of the top-down cascade pathway via element-wise addition leads to the performance degradation without any parameter loss.
A simple top-down addition strategy on the early layer of DCMNet reinforces the feature connections of all the following layers, and leads to the performance improvement in depth estimation.
Also, PPM head and dense concatenation improve the performance across all the metrics with additional model parameters, and our method with all the components combined together achieves the best results.

We compare our SwinDepth to recent hierarchical Transformers for dense prediction.
SwinDepth produces depth estimation results on par with state-of-the-art architectures with smaller parameters (DPT \cite{ranftl2021vision} (123M), HRFormer \cite{yuan2021hrformer} (56.0M)).

\section{Conclusion}
In this paper, we present that the convolution-free hierarchical Transformer outperforms existing ConvNet-based architectures in monocular depth estimation. 
Also, we propose DCMNet to improve the model performance by reinforcing the interconnection of the decoding layers with top-down addition and dense concatenation. 
We achieve state-of-the-art performance compared to competitive unsupervised algorithms using monocular sequences on the KITTI and Make3D dataset.

\section{Acknowledgement}
This work was supported by Institute of Information \& communications Technology Planning \& Evaluation (IITP) grant funded by the Korea government(MSIT) [NO.2021-0-01343, Artificial Intelligence Graduate School Program (Seoul National University)]
\bibliographystyle{IEEEtran}
\bibliography{root.bib}

\end{document}